\documentclass[runningheads]{llncs}
\usepackage{graphicx}
\usepackage{amsmath}
\usepackage{amssymb}
\usepackage[T1]{fontenc}

\graphicspath{{./figures/}}

\begin{document}

\title{Sampling Strategies for Mitigating Bias in Face Synthesis Methods}

\author{Emmanouil Maragkoudakis\inst{1}
\and
Symeon Papadopoulos\inst{2}
\and
Iraklis Varlamis\inst{1}
\and
Christos Diou\inst{1}
}
\authorrunning{Maragkoudakis et al.}

\institute{Department of Informatics and Telematics\\ Harokopio University of Athens, Athens, Greece \and
Information Technologies Institute\\ Centre for Research and Technology Hellas, Thessaloniki, Greece}

\maketitle            

\begin{abstract}
  Synthetically generated images can be used to create media content or to complement datasets for training image analysis models. Several methods have recently been proposed for the synthesis of high-fidelity face images; however, the potential biases introduced by such methods have not been sufficiently addressed. This paper examines the bias introduced by the widely popular StyleGAN2 generative model trained on the Flickr Faces HQ dataset and proposes two sampling strategies to balance the representation of selected attributes in the generated face images.
  We focus on two protected attributes, gender and age, and reveal that biases arise in the distribution of randomly sampled images against very young and very old age groups, as well as against female faces. These biases are also assessed for different image quality levels based on the GIQA score. To mitigate bias, we propose two alternative methods for sampling on selected lines or spheres of the latent space to increase the number of generated samples from the under-represented classes.
  The experimental results show a decrease in bias against underrepresented groups and a more uniform distribution of the protected features at different levels of image quality.

\end{abstract}

\section{Introduction}
High-fidelity face synthesis is useful for several applications, such as new content creation \cite{Moschoglou2020-ry,Shen2018mf}, use of synthetic faces with 3D models to train face identification algorithms \cite{marriott20213d,3dganColorization2022}, gaze estimation \cite{Fischer_2018_ECCV,Kim2020-en} and others \cite{antispoofingYang2022,Wu2019-ha}. Face synthesis using Generative Adversarial Networks (GANs) has been an intensively investigated research area in the last few years \cite{YE2019294,ruiz2020morphgan,trgan2020}. Initially proposed  in \cite{goodfellow2020generative}, GANs consist of two neural networks that contest with each other with the aim to produce synthetic samples drawn from the training data distribution. GAN-based models are now able to generate high-quality images of human faces, with StyleGAN \cite{StyleGAN2} being one of the most popular family of methods in this category.

The synthetic images generated by such methods tend to follow the same distribution of facial attributes (e.g., skin color) as their training dataset, and may therefore reflect the biases of the data collection process. Unless appropriate mitigation measures are taken, bias can be transferred or even amplified in the generated results. Depending on the application, such biases can have negative real-world consequences (e.g., reproduction of stereotypes on synthetic content, training of biased models for face analysis). On the opposite side, the controlled generation of synthetic images could actually help remove biases that exist in real image datasets and can help train unbiased models, e.g., by oversampling attributes that are initially missing or are underrepresented.

The objective of this work is to measure bias in GAN-based face synthesis methods and in StyleGAN2 in particular, as well as to propose methods that can help reduce this bias, by using latent space sampling strategies. We focus on two protected attributes, namely age, and gender, and initially assess bias in a pre-trained StyleGAN2 model. To this end, we use pre-trained age and gender estimation models and evaluate the distribution of these protected attributes in the resulting dataset. Our analysis shows that the model is indeed biased towards certain age groups and towards female faces. Bias is also associated with the generated image quality since it seems to be amplified in high-quality images. To mitigate this bias, we propose two post-processing sampling strategies in the GAN latent space.

In summary, the main contributions of this work are the following:
\begin{itemize}
    \item An investigation of StyleGAN-based face synthesis methods with respect to the handling of bias introduced in protected features such as age and gender, also taking into account the image quality dimension.
    
    \item A bias mitigation method that takes as input a pair of selected face images with different attribute values, and generates intermediate images in an attempt to increase the representation of certain sample groups.

    \item A second mitigation method that performs a series of samplings on multivariate normal distributions around a single image from the original dataset, which belongs to an underrepresented class. The method results in increased representation of the input image attributes in the final dataset.
\end{itemize}

\begin{figure*}
    \begin{tabular}{ccc}
    \includegraphics[width=.5\linewidth]{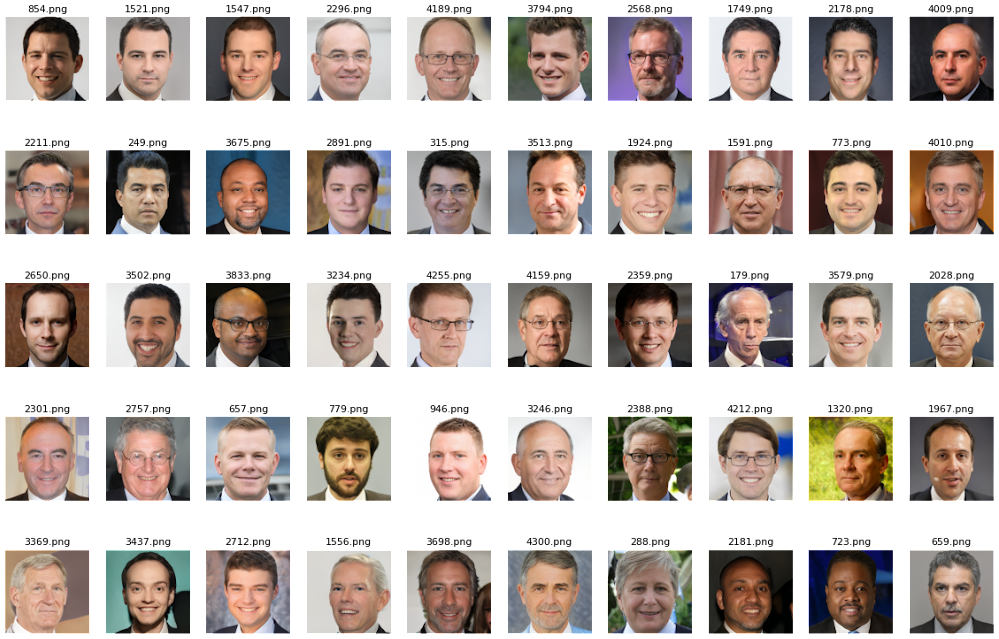} & &
    \includegraphics[width=.5\linewidth]{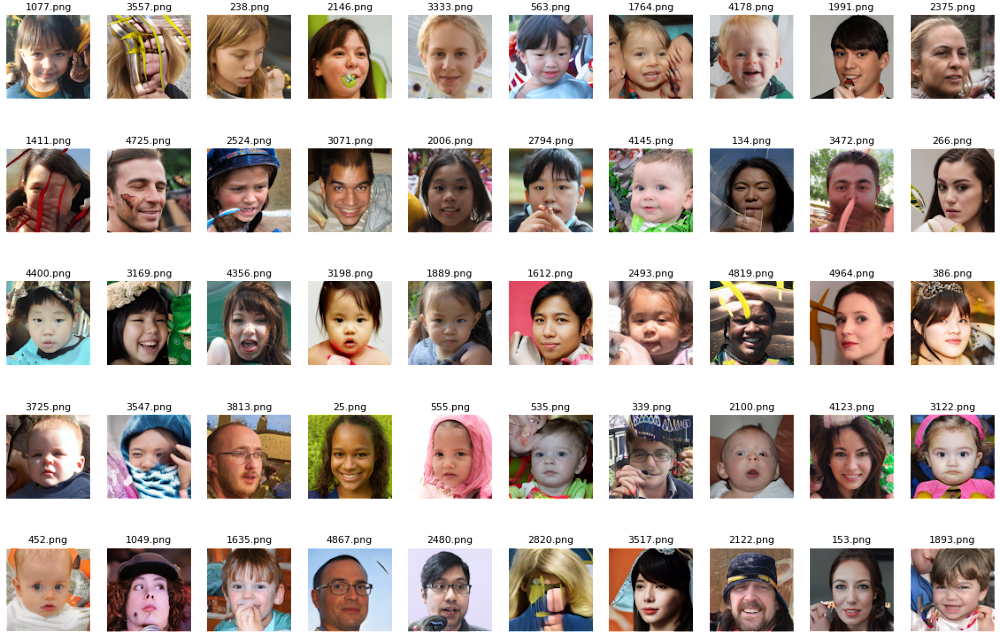} \\
    (a) & & (b) 
    \end{tabular}
    \caption{Face images generated by a pre-trained StyleGAN2 model. The left figure (a) illustrates the highest quality images in a dataset of 5000 images, as measured by the GIQA score. The right figure (b) shows the worst quality images of the same dataset. There is a pronounced gender and age bias in the two image quality groups as we can see that the majority of the top 50 images are faces of white adult men while the majority of the bottom 50 are images of children of color.}
    \label{fig:top_bottom}
\end{figure*}

\section{Related work}
\label{relwork}

\textit{Visual bias} in image synthesis methods can be defined as the tendency of generative algorithms to carry the imbalance and under-representation of certain attributes found in the dataset that has been used for their training. This kind of bias may result in distortion of the representations of certain attributes, such as age, gender, or skin color. Wang et al. in \cite{REVISE} describe visual bias in models as a result of datasets that partially capture the visual world and represent only a particular distribution of visual data. 
Fabbrizzi et al. \cite{fabbrizzi2022survey} refer to this bias as \textit{selection bias}, and define it as any disparities or associations created by the process that includes subjects in the visual training dataset.

In the domain of image generation, \textit{de-biasing} refers to the act of mitigating or eliminating biases in generative algorithms. These biases can lead to inaccurate or distorted representations of certain characteristics such as age, gender, or skin color. De-biasing is the attempt to produce synthesized images while reducing discriminatory representations and ensuring fairness and accuracy in image synthesis applications. 
De-biasing can take place before model training when preparing the data, during the model training process, or after the model training is complete. Ntoutsi et al. \cite{ntoutsi2020bias} refer to these three cases as \textit{pre-processing}, \textit{in-processing}, and \textit{post-processing} respectively. Pre-processing involves the preparation of the training data by fixing imbalance and under-representation of classes. In-processing refers to the application of de-biasing techniques during the training of the model. Post-processing is the process of applying de-biasing techniques after the training is complete, directly on the model outputs.

The issue of bias has been of great concern to the scientific community. According to Pessach et al. in \cite{AlgoFairness}, an ever-increasing number of decisions are guided by Artificial Intelligence algorithms, and the presence of bias in these algorithms can significantly affect human lives. In their article, they point out that a major challenge in the effort to eliminate algorithmic bias stems from bias in the training data. They state that this challenge can arise, for example, if there are underrepresented populations in the datasets.

Despite the interest of the research community in fairness and mitigation of bias in classification tasks \cite{mehrabi2021survey,ntoutsi2020bias}, the problem of fairness and diversity in synthetic imagery has received much less attention  \cite{FairStyle,xu2018fairgan}. Most research on the bias and fairness of generative models such as GANs seeks either to eliminate the negative impact of using imbalanced data on the generated results or to detect and justify the presence of biases.

Previous research on assessing the adequate representation of different visual attributes in datasets has led to the creation of tools such as REVISE (REvealing VIsual biaSEs) \cite{REVISE}, which facilitates the investigation of a set of images for potential bias based on the diversity of representation of different objects and features. As the tool identifies biases of various causes, it suggests actions that can be taken to mitigate any bias detected by the user.

Georgopoulos et al. in \cite{georgopoulos2021mitigating} condition the face generation on discriminative representations for each attribute instead of collapsing attribute information into a single label. They propose to transfer the joint demographic style of each sub-population in the generated images and in that way modify the attributes of each face and enhance the diversity of the dataset. To combine the different representations capturing complex facial patterns related to each attribute, they propose a novel extension to AdaIN \cite{Huang2017-zr}, tailored to handle the mixing of multiple attribute representations by introducing a tensor-based mixing structure that captures multiplicative attribute interactions in a multi-linear fashion.

Maluleke et al. \cite{maluleke2022studying} explore the impact of racial representation in training datasets and conclude that the trained models retain the bias derived from the training dataset. The same study notices a correlation between the observable quality of images and the representation of race in the distribution that generated them.
According to the findings of the study, StyleGAN2-ADA closely matches the racial composition of the corresponding training data.

Doubinsky et al. in \cite{doubinsky2022multi} address the issue of attribute entanglement in GANs. First, they observed high correlations between features in the generated data, so they represented the distribution of binary attributes of a generated dataset as an $m$-dimensional contingency table. As a result, they propose a simple method of oversampling to generate a dataset $S$ and then sub-sample it to generate a dataset $B$, the contingency table of which is expected to be balanced. 

Ramaswamy et al. in \cite{ramaswamy2021fair} proposed a pre-processing method of constructing a de-biased dataset consisting of both real images used for training a generator and synthetic images generated by sampling from perturbing vectors in the latent space of the trained generator. After constructing the de-biased dataset consisting of both real and synthetic images, the authors retrained the generator to minimize the observed bias in the data.

Another regularization method was proposed by Tartaglione et al. in \cite{tartaglione2021end} as well. Their strategy entangles features extracted from the same target class while disentangling biased features at training time by incorporating regulations on the loss function used for the training of the model.
Their proposed methods extend to the subject of mitigating a known bias towards some protected characteristic, such as gender or age group, in a set of synthetic images and classify as post-processing methods as they treat the pre-trained model as a black box and only act upon its input, to modify the distribution of the resulting synthetic images.

\section {Sampling strategies for bias mitigation}

Based on the findings of the related work survey, it is clear that the fairness of GAN-based face synthesis methods can be significantly affected by bias in the training dataset. Let $\mathcal{G}$ be a face synthesis model and $\mathcal{A}$ a face attribute (e.g., gender, age, skin color), which can take values $a_1, \dotsc, a_n$. To generate a face image using GAN-based face synthesis models, one randomly selects a sample $z \in \mathbb{R}^d$ in the latent space to produce an output face image, $f$. If $\mathcal{G}$ is biased with respect to $a$, then uniform random sampling in the latent space will lead to a non-uniform distribution of the values $a_i$, $i = 1, \dotsc, n$ in the resulting dataset. If the bias is significant, then some values of $\mathcal{A}$ will be prevalent, while others will be underrepresented.

Another, less obvious way in which bias manifests in the image synthesis process, is via image quality. Let $q \in [0, 1]$ be a value quantifying the quality of the generated images, according to criteria such as distortions, blurring, artifacts, etc. \cite{GIQA}. It may then be possible that the bias (i.e., deviation from the uniform distribution of $\mathcal{A}$) is further amplified if we condition on specific ranges of $q$. For example, the imbalance may be more pronounced for high-quality images, i.e., $\mathcal{A}|q > 0.5$. An instance of this problem is shown clearly in Figure \ref{fig:top_bottom}, which illustrates quality-dependent bias of a pre-trained StyleGAN2 model.

One approach to addressing these problems is to sample the latent space in a non-uniform way, oversampling the under-represented attribute values while undersampling the over-represented ones. This approach belongs to the class of ``post-processing'' bias mitigation strategies, since the goal is to counter the bias inherent in an already trained $\mathcal{G}$. Through appropriate sampling in the latent space, $z$, one can produce a dataset with uniformly distributed values of the attribute $\mathcal{A}$. Note that the same ideas can be extended to more than one attributes. For example, one may be interested in a pair of attributes $(\mathcal{A}_1, \mathcal{A}_2)$, with values $(a_{1,i}, a_{2, j})$, $i=1,\dotsc, n_1$, $j=1,\dotsc,n_2$. The benefit of the proposed approach is that it does not require retraining the generative model to mitigate bias. The following two subsections describe two sampling strategies for implementing this approach. All visual examples correspond to a StyleGAN2 model \cite{StyleGAN2} that has been trained on the Flickr Faces HQ (FFHQ) dataset.

\subsection{Bias and fairness}
To evaluate the reduction of bias through the proposed methods, we must first define a bias quantification method.

Our debiasing strategy for GAN models focuses on quantifying and mitigating bias through the use of three key metrics: imbalance ratio, imbalance degree, and log-likelihood index. In the context of GAN models, we define bias as the occurrence of large discrepancies or imbalances in these metrics. 

The imbalance ratio quantifies the degree of imbalance in a dataset's categorical attributes. It is defined as the ratio of the greatest and least number of unique attribute values. A higher ratio shows a greater discrepancy between the most and least frequent attribute values, indicating a greater degree of imbalance. A completely balanced attribute has an imbalance ratio of 1, indicating that all attribute values have the same count. The imbalance ratio (\text{{IR}}) is calculated as:

\[ IR = \frac{{\max(\text{{counts}})}}{{\min(\text{{counts}})}} \]

\noindent where \(\max(\text{{counts}})\) represents the maximum count among the unique attribute values, and \(\min(\text{{counts}})\) represents the minimum count.

The imbalance degree (ID) \cite{ORTIGOSAHERNANDEZ201732} is used to quantify the degree of class imbalance in scenarios involving binary and multi-class classification. The difference between a balanced distribution and the actual imbalanced problem is represented by condensing the information about the class distribution into a single numerical value that reflects the dataset's imbalance as well as the number of majority and minority classes. In a perfectly balanced scenario, ID would be equal to zero. The imbalance degree (\text{{ID}}) is calculated as:

\[ID = \frac{H(p, b)}{H(p_m, b)} + (m - 1)\]

\noindent where \text{{H(p,q)}} is the Hellinger distance between two probability distributions \text{{p}} and \text{{q}} and is calculated as follows:

\[H(p, q) = \frac{1}{\sqrt{2}} \sqrt{\sum_{i=1}^{c} \left(\sqrt{p_i} - \sqrt{q_i}\right)^2} \]

\noindent where $p_i$ and $q_i$ are the probabilities of each category in the distributions $p$ and $q$, respectively. In the context of ID, $p$ represents the observed distribution of categories in the attribute, and $q$ represents the uniform distribution.

The log-likelihood index (LLI) measures the likelihood of a model generating images of the minority class, with higher values indicating better results. The log-likelihood index (LLI) is calculated as:

\[LLI = 2 \sum_{i=1}^{c} \left(p_i \log(p_i \cdot c)\right)\]

\noindent where $c$ is the total number of categories in the attribute and $p_i$ 
  is the probability of category $i$ in the observed distribution.

\subsection{Line sampling}
The first strategy assumes a line in the latent space of image representation that connects two images of the generated dataset.
An effective selection strategy for the input images is to pick two images from the under-represented class when you want to balance out the dataset or one from each class when you just want to expand the size of the dataset. This way we can produce intermediate images between the two selected images to perform the sampling.

This method allows control over the sampling process with the newly sampled images ranging between the pairs of selected images, but it also comes with the drawback of having a less diverse dataset since the new images will bear similarity to images in the generated dataset.

The method operates in two phases. In the initial phase, we focus on mitigating gender imbalance within each age group for the top 25\% and middle 50\% of images based on their GIQA quality scores. By excluding the bottom 25\% we ensure that images of high quality are being selected for the re-sampling during the balancing process.

Within the selected high-quality images we access the gender distribution and identify the under-represented gender within each age group. After identifying the under-represented gender, the method selects enough pairs of images of the selected gender within the age group to perform the line sampling procedure. The goal of the first phase is to balance out the gender attribute for each age group. Afterwards the age and gender of the new images, as well as their quality score is calculated to be used for the second sampling.

Once the initial gender balancing within age groups is completed, we move to the second phase, which focuses on balancing the age groups themselves. The method now calculates the average number of images per age group in the top and middle quality levels based on the results of the fist phase. For each quality level that contains less images than the average, the method samples pairs of images of the age group, one of each gender and performs the line sampling procedure again to reach the desired average number of images per age group.  

To implement the line sampling procedure, the method uses the latent vectors that produced the selected images of each pair as the edges for the line samplings. We can then sample using different mixing levels between these vectors. More specifically, we create a direction vector by subtracting between the latent space vectors corresponding to the images, i.e.,
\begin{displaymath}
   v = z_{end} - z_{start}
\end{displaymath}
\noindent where $v$ is the direction vector, $z_{start}$ is the vector used to construct the first image and $z_{end}$ is the vector used to construct the second image.

\begin{figure}
  \centering
  \includegraphics[width=\linewidth]{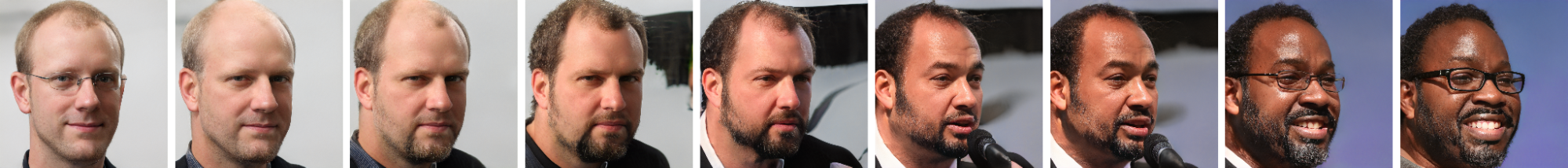}
  \caption{Moving along a straight line of the latent space. The method has selected two images with the same age and gender attributes to perform the linear sampling. In this scenario we are looking for more images of middle aged men to balance out a dataset that contains more images of women in the corresponding age group. }
  \label{fig:linesampling}
\end{figure}

Having calculated the direction vector, we can now move on to the line segment by adding to the original image the direction vector multiplied by a step value ranging between 0 and 1. The step values allow the recreation of several intermediate images between the initial and the final image.
The images are produced using the following formula:
\begin{displaymath}
z_i = z_{start} + v \cdot  c_i, c_i \in [0,1]
\end{displaymath}
\noindent where $z_i$ is an intermediate image and $c_i$ is a step value between 0 and 1. Figure \ref{fig:linesampling} illustrates an example of this sampling method

\subsection{Sphere sampling}
The second proposed strategy is called sphere sampling because it samples images around a starting image.
It performs sampling based on a multivariate normal distribution around selected images (used as means of the distribution) of the generated dataset. The samples generated with this method are still related to the selected seed image but are more diverse than the ones created by the first method since multiple seeds are used. 

The method operates in two phases in the same way as the line sampling method. The only difference is that it samples around one image's latent vector each time, so the selected images are not in pairs. In the first phase this method samples around images of the under-represented class with the same goal of balancing the gender attribute for each age group. In the second phase the sampling procedure takes place with images of both genders in an effort to keep the age group balanced according to gender while trying to reach the desired average number of images per age group.

The resulting enriched dataset contains new images that fall into the classes of the selected seed images and add diversity to the dataset mitigating the previously existing bias in favor of the over-represented class.

\begin{figure}
  \centering
  \includegraphics[width=\linewidth]{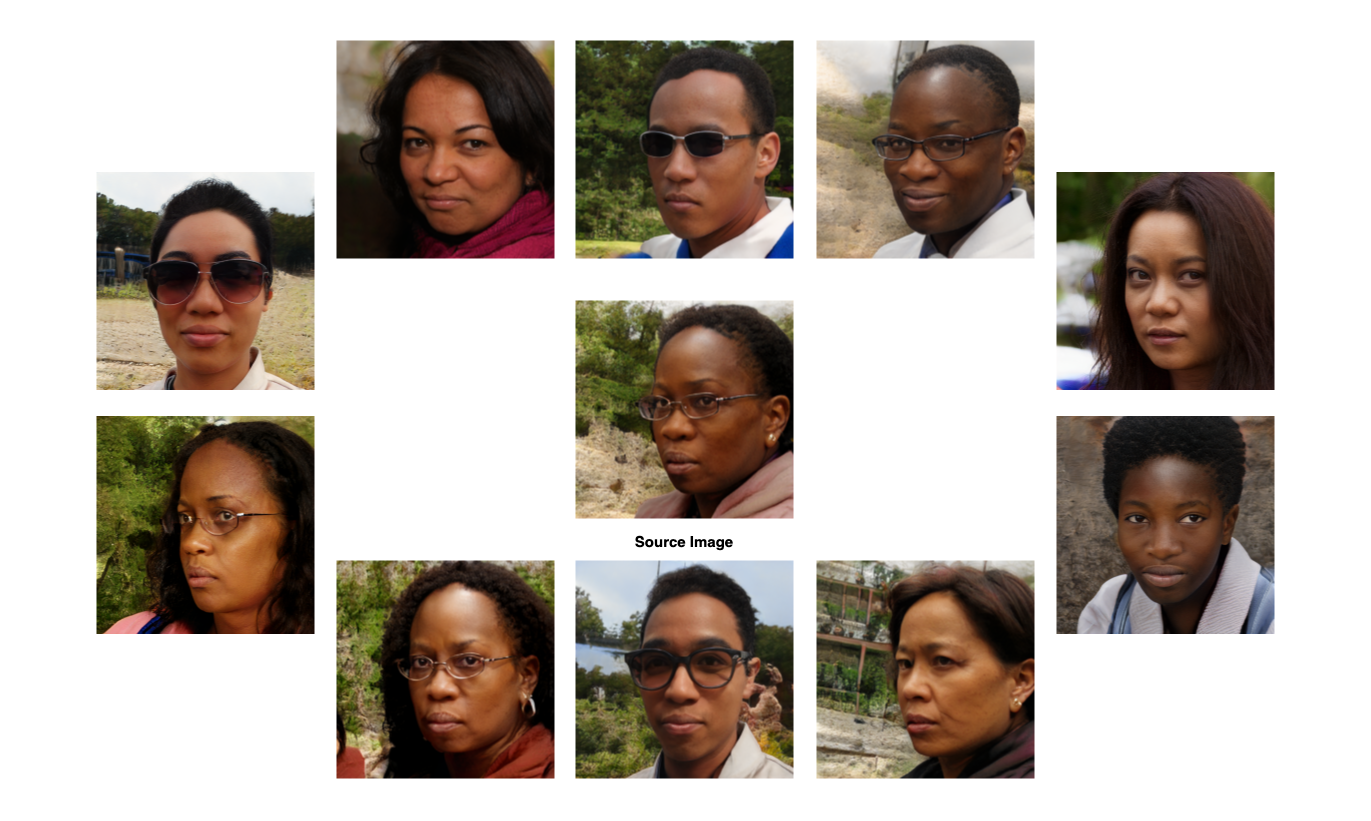}
  \caption{Moving in a sphere around a selected point in the latent space.}
  \label{fig:sphere_sampling}
\end{figure}

The images are produced by sampling randomly on multivariate normal distributions of this form:

\begin{displaymath}
X \sim N_{d}(z_{seed},C_v)
\end{displaymath}
\noindent where $z_{seed}$ the vector of the selected seed image and $C_z$ a diagonal covariance matrix $0.1\cdot I$. Figure \ref{fig:sphere_sampling} illustrates an example using this method.

\section{Experiments}
\label{experiments}
\subsection{Assessing the bias of a pre-trained StyleGAN2 model}

We used a StyleGAN2 model trained on the FFHQ dataset to generate a set of 5,000 synthetic images by selecting points $z$ in the latent space uniformly at random. Age and gender are used as the protected attributes and evaluate the diversity and bias of the generated dataset for a variety of image quality levels. The age attribute is split into four age categories (without loss of generality). The protected attributes were estimated using an Age-Gender-Estimation image classification model \cite{AgeGender} trained on the IMDB-WIKI dataset \cite{Rothe-IJCV-2018}.  

The image quality assessment method used was the GIQA algorithm \cite{GIQA}, which produces a ranking of the collection of synthetic images in decreasing order of quality.

To quantify bias, we calculate the three metrics mentioned above: \textit{imbalance ratio}, \textit{imbalance degree}, and \textit{log likelihood index}.

\begin{figure*}[!htb]
  \centering
  \begin{tabular}{c}
    \includegraphics[width=.9\linewidth]{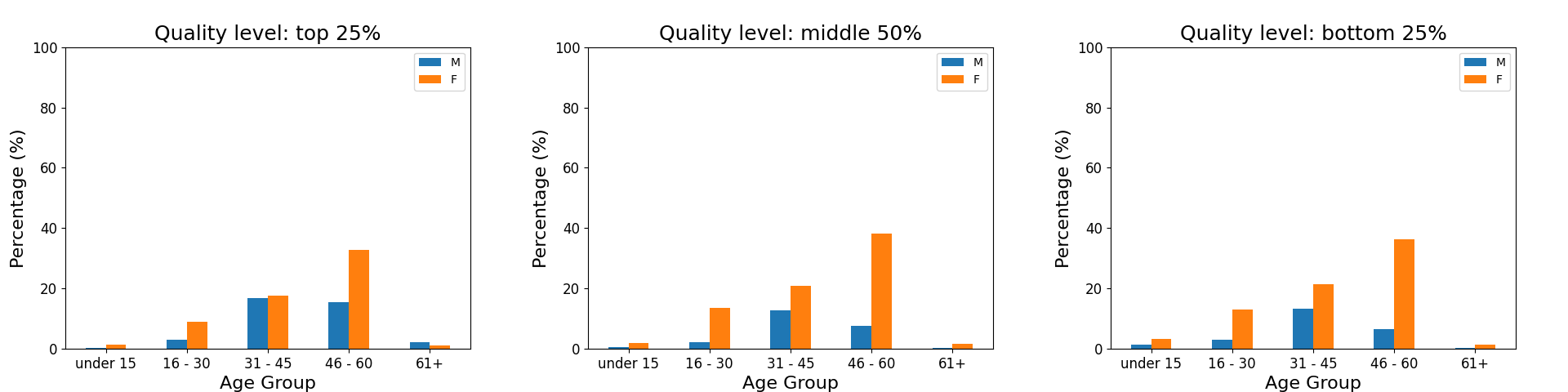}\\
    (a) initial dataset\\
    \includegraphics[width=.9\linewidth]{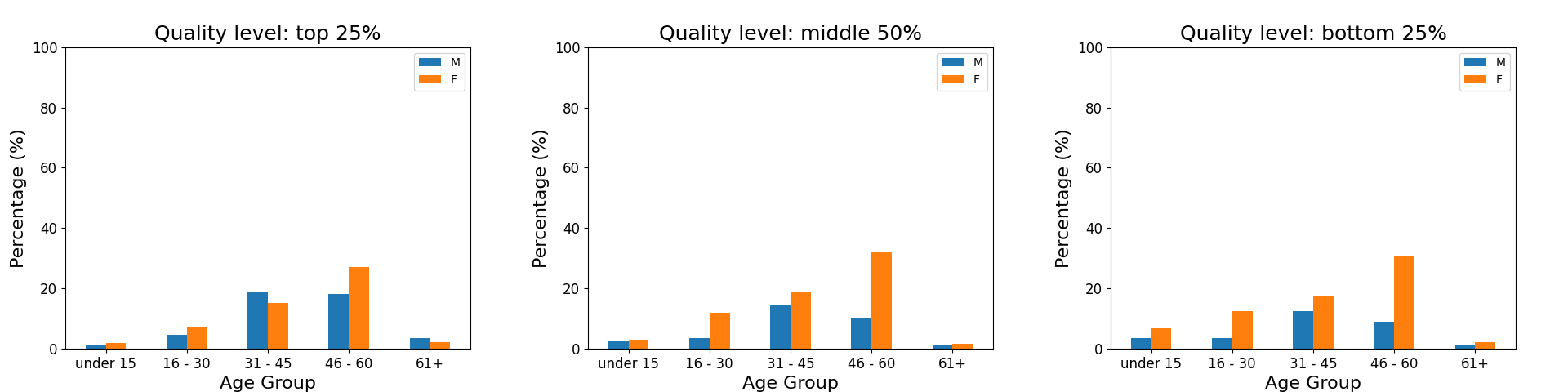}\\
    (b) line sampling\\
     \includegraphics[width=.9\linewidth]{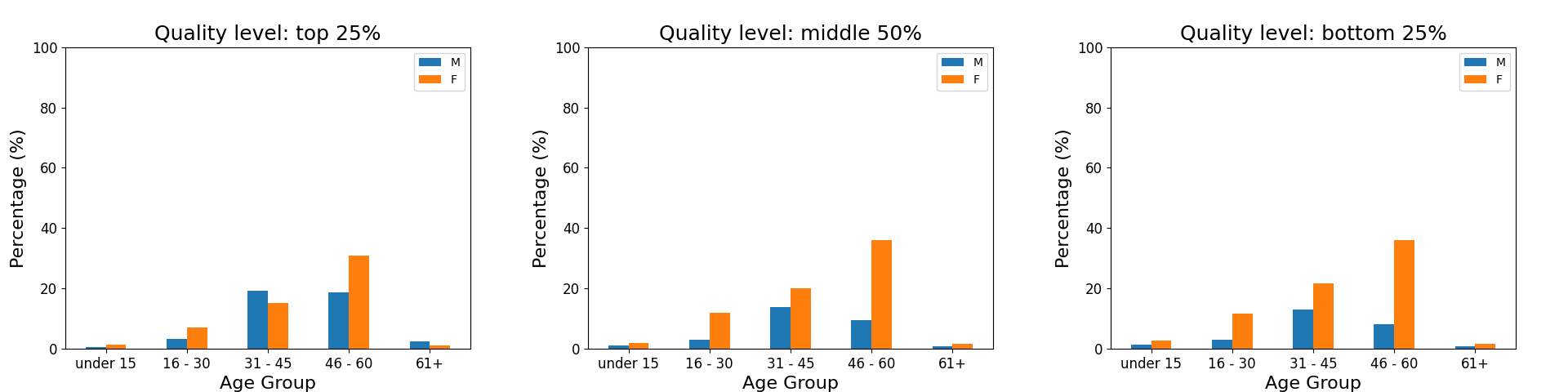}\\
     (c) sphere sampling
    \end{tabular}
  \caption{Plot representation of the distribution of features in the initial dataset.}
  \label{fig:distplots}
\end{figure*}

Figure \ref{fig:distplots}(a) illustrates the distribution of the age and gender in the original dataset, generated via uniform random sampling, while the first rows of Table \ref{tab:metrics} present the corresponding metrics. Based on these results, the model demonstrates a strong age bias with respect to the selected age groups, with an imbalance ratio of 21.33. The imbalance ratio (majority/minority class) for gender is 2.63 (values closer to 1 denote better balance).
\begin{table*}
\caption{Metrics for the original dataset and the dataset resulting after application of the two sampling strategies}
\label{tab:metrics}
\begin{tabular}{|l|l|l|l|}
\hline
Selected Attribute & Imbalance Ratio ($\downarrow$) & Imbalance Degree ($\downarrow$) & Log Likelihood Index ($\uparrow$) \\
\hline
\multicolumn{4}{|c|}{Original dataset}\\
\hline
Age group & 21.330 & 3.654 & 0.826 \\
\hline
Gender & 2.631 & 0.303 & 0.209 \\
\hline
\multicolumn{4}{|c|}{Dataset after line sampling}\\
\hline
Age group & 11.392 & 3.594 & 0.604 \\
\hline
Gender & 1.837 & 0.195 & 0.088 \\
\hline
\multicolumn{4}{|c|}{Dataset after sphere sampling}\\
\hline
Age group & 16.709 & 3.648 & 0.817 \\
\hline
Gender & 2.146 & 0.243 & 0.136 \\
\hline
\end{tabular}
\end{table*}

For the evaluation of the line sampling method, pairs of images with different age/gender attributes were selected and line sampling was applied until sufficient 
images was produced to balance the ``Gender'' attribute. For this experiment, 93 pairs of images were selected automatically.

\begin{figure}[!htb]
  \centering
  \begin{tabular}{c}
  \includegraphics[width=\linewidth]{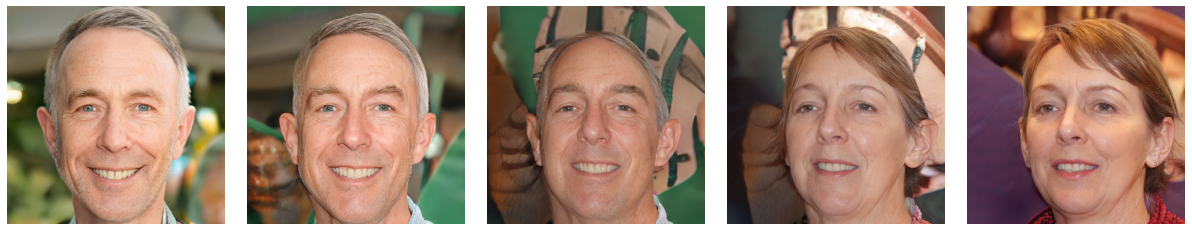}\\
  (a) White middle-aged man to a white middle-aged woman\\
   \includegraphics[width=\linewidth]{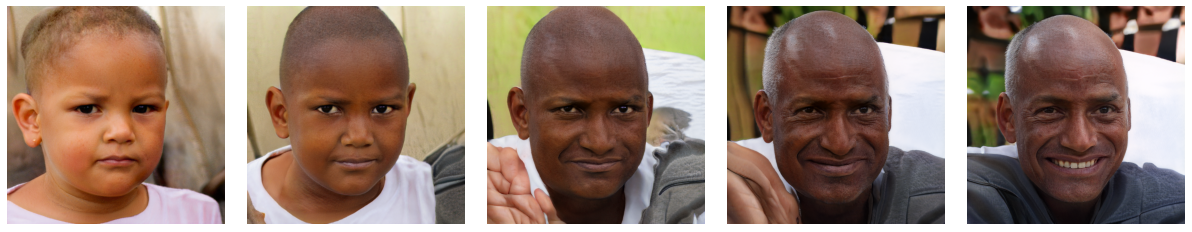}\\
  (b) Black baby boy to a black old man\\
  \end{tabular}
  \caption{Two examples of the line sampling strategy}
  \label{fig:line_sampling_examples}
\end{figure}

Each line segment was used to generate 37 samples, resulting in a total of 3441 new synthetic face images that enriched the initial data set. After removing any resulting duplicate samples in the latent space, the images were reduced to 3394.
Indicative examples of line sampling being performed are the transitions shown in Figure \ref{fig:line_sampling_examples}.

After performing the same evaluation that was used on the original set of images produced with the StyleGAN2 algorithm, we get the distributions presented in the plots of Figure \ref{fig:distplots}(b). As seen in Table 
\ref{tab:metrics}, the imbalance ratio for gender is 1.83 which denotes a great improvement in balance, while the age imbalance has also been significantly reduced. Interestingly, the imbalance of the produced images in terms of quality remains the same.

The plots of Figure \ref{fig:distplots}(b) show that the distribution of gender in relation to image quality has changed significantly. Especially for the category of high-quality images, we notice that the number of images of women and men are equal. Moreover, there is an increase in the number of images of children and older adults which were very few in the high-quality category.

In addition to the distribution of gender, the distribution of images in the various age groups has also been largely balanced. The standard deviation of the percentages of images belonging to each age group on the images of the highest quality before applying the method is 18.40 while after applying the method the standard deviation is 16.77.

\subsection{Sphere sampling method evaluation}

For the evaluation of the sphere sampling method, 86 images were selected as seeds for sampling on the multivariate normal distribution, again with the aim to balance the ``Gender'' attribute, as in line sampling. From each distribution, 37 samples were selected producing a total of 3182 additional synthetic face images to enrich the initial data set. After removing some duplicates and similar images the new images were reduced to 3160.

After performing once again the same statistical control that was used on the original set of images produced with the StyleGAN2 algorithm and on the previous method, we get the results depicted in Figure \ref{fig:distplots}(c). The imbalance ratio for gender, in this case, is 2.14 that indicates that there is still room for improvement when compared to the line sampling method. 

The distribution of the gender attribute in relation to quality has improved significantly in comparison to the results of the model prior to the introduction of the sphere sampling method, as seen in the table and the plots. We notice that as with line sampling, the distribution has changed drastically, especially in the category of high-quality images with the number of images of women now balanced against the number of men. Looking at the dataset after adding the images generated by the method, it is obvious that the new images are more diverse and less dependent on the seed images compared to the ones generated using the previous method. 

This method provides images that share common characteristics with the seed images of the sampling but introduces a level of diversity
that was not possible with the line-based sampling method. 

In addition to the distribution of gender, the distribution of images in the various age groups has also been largely balanced. The standard deviation of the percentages of images belonging to each age group on the images of the highest quality before applying the method is 18.40 while after applying the method the standard deviation is 17.42.

\subsection{Discussion}

In comparison to the original dataset, the results from the two proposed bias mitigation methods showed promising improvements. The Line sampling method successfully reduced the imbalance ratio and imbalance degree for both the ``Age group'' and ``Gender'' attributes. The imbalance ratio was lowered to 11.392 with an imbalance degree of 3.594 for ``Age group'', compared to the original dataset imbalance ratio of 21.330 and imbalance degree of 3.654. In the case of ``Gender'', the imbalance ratio was reduced to 1.837 and the imbalance degree to 0.196, compared to the original dataset imbalance ratio of 2.631 and imbalance degree of 0.304. Similarly, the sphere sampling method demonstrated improvements for both ``Age group'' (a reduced imbalance ratio of 16.709 with a imbalance degree of 3.648) and ``Gender'' (a reduced imbalance ratio of 2.146 and imbalance degree of 0.243). 

When compared to Sphere sampling, Line sampling seems to achieve slightly better results in terms of reducing imbalance. However, Sphere sampling exhibits slightly greater log likelihood index values compared to line sampling. While the log likelihood index values for both methods were relatively low, indicating potential room for improvement, these results indicate that the proposed sampling methods have the potential to effectively mitigate bias in the dataset for multiple attributes.

Each method has distinct strengths and limitations that must be taken into account in real-world applications. Line sampling is a controlled method for sampling along a linear path that connects two selected synthetic images. Our results show that although this strategy works well in reducing bias, it may result in a large number of visually similar photos, a typically undesired property for an image collection. 
Sphere sampling, on the other hand, is a random sampling method that chooses vectors in the latent space vicinity around a vector corresponding to an image. Our findings indicate that this strategy is similarly successful in reducing bias in face synthesis models. However, Sphere sampling may not be appropriate for all use cases, particularly when fine-grained control over the sampling process is required.

Future research could look into ways to automate image selection for the re-sampling process as well as non-linear transitions between two images to improve the effectiveness of Line sampling. Such approaches could result in more efficient bias reduction while retaining image diversity.

Furthermore, future studies could look into strategies for improving both the Line and the Sphere sampling approaches, their combination, as well as its joint application with other techniques, like prepossessing of the training dataset, to produce improved outcomes in minimizing bias in face synthesis models. This could include experimenting with different sampling methods or constructing advanced algorithms to control the sampling process.

\section{Conclusions}
\label{sec:conclusions}
In this study, we proposed and evaluated two methods for reducing bias in face synthesis models: Line and Sphere sampling. Our results indicate that both strategies effectively mitigate bias in the collection of generated images, but each method has unique strengths and limitations that must be considered before applying in practical settings.

Bias in AI is an important issue that raises ethical concerns, particularly in applications with sensitive data such as medical or criminal records. The suggested Line sampling and Sphere sampling approaches make notable contributions to the field of bias mitigation in face synthesis models, which is an important step in developing fair and less discriminatory AI systems.

These strategies could be highly effective for decreasing bias and enhancing image variety. Further research can build upon our findings to explore new ways to improve bias mitigation techniques and achieve even better results in face synthesis models.

\section*{Acknowledgements}

The work leading to these results has received funding from the European Union’s Horizon 2020 project REBECCA  under Grant Agreement No. 965231 and Horizon Europe project MAMMOth under Grant Agreement No. 101070285. 

\bibliographystyle{splncs04}
\bibliography{biblio}

\begin{thebibliography}{10}
\providecommand{\url}[1]{\texttt{#1}}
\providecommand{\urlprefix}{URL }
\providecommand{\doi}[1]{https://doi.org/#1}

\bibitem{doubinsky2022multi}
Doubinsky, P., Audebert, N., Crucianu, M., Le~Borgne, H.: Multi-attribute balanced sampling for disentangled gan controls. Pattern Recognition Letters  \textbf{162},  56--62 (2022)

\bibitem{fabbrizzi2022survey}
Fabbrizzi, S., Papadopoulos, S., Ntoutsi, E., Kompatsiaris, I.: A survey on bias in visual datasets. Computer Vision and Image Understanding p. 103552 (2022)

\bibitem{Fischer_2018_ECCV}
Fischer, T., Chang, H.J., Demiris, Y.: Rt-gene: Real-time eye gaze estimation in natural environments. In: Proceedings of the European Conference on Computer Vision (ECCV) (September 2018)

\bibitem{georgopoulos2021mitigating}
Georgopoulos, M., Oldfield, J., Nicolaou, M.A., Panagakis, Y., Pantic, M.: Mitigating demographic bias in facial datasets with style-based multi-attribute transfer. International Journal of Computer Vision  \textbf{129}(7),  2288--2307 (2021)

\bibitem{goodfellow2020generative}
Goodfellow, I., Pouget-Abadie, J., Mirza, M., Xu, B., Warde-Farley, D., Ozair, S., Courville, A., Bengio, Y.: Generative adversarial networks. Communications of the ACM  \textbf{63}(11),  139--144 (2020)

\bibitem{GIQA}
Gu, S., Bao, J., Chen, D., Wen, F.: Giqa: Generated image quality assessment. In: Computer Vision--ECCV 2020: 16th European Conference, Glasgow, UK, August 23--28, 2020, Proceedings, Part XI 16. pp. 369--385. Springer (2020)

\bibitem{Huang2017-zr}
Huang, X., Belongie, S.: Arbitrary style transfer in real-time with adaptive instance normalization. In: Proceedings of the IEEE international conference on computer vision. pp. 1501--1510 (2017)

\bibitem{FairStyle}
Karakas, C.E., Dirik, A., Yal{\c{c}}{\i}nkaya, E., Yanardag, P.: Fairstyle: Debiasing stylegan2 with style channel manipulations. In: European Conference on Computer Vision. pp. 570--586. Springer (2022)

\bibitem{StyleGAN2}
Karras, T., Laine, S., Aittala, M., Hellsten, J., Lehtinen, J., Aila, T.: Analyzing and improving the image quality of stylegan. In: Proceedings of the IEEE/CVF conference on computer vision and pattern recognition. pp. 8110--8119 (2020)

\bibitem{trgan2020}
Kezebou, L., Oludare, V., Panetta, K., Agaian, S.: {TR-GAN: thermal to RGB face synthesis with generative adversarial network for cross-modal face recognition}. In: Agaian, S.S., Asari, V.K., DelMarco, S.P., Jassim, S.A. (eds.) Mobile Multimedia/Image Processing, Security, and Applications 2020. vol. 11399, p. 113990P. International Society for Optics and Photonics, SPIE (2020)

\bibitem{3dganColorization2022}
Khan, Z., Umar, A.I., Shirazi, S.H., Shahzad, M., Assam, M., El-Wakad, M.T.I.M., Attia, E.A.: Face recognition via multi-level 3d-gan colorization. IEEE Access  \textbf{10},  133078--133094 (2022)

\bibitem{Kim2020-en}
Kim, J.H., Jeong, J.W.: Gaze in the dark: Gaze estimation in a low-light environment with generative adversarial networks. Sensors (Basel)  \textbf{20}(17), ~4935 (2020)

\bibitem{maluleke2022studying}
Maluleke, V.H., Thakkar, N., Brooks, T., Weber, E., Darrell, T., Efros, A.A., Kanazawa, A., Guillory, D.: Studying bias in gans through the lens of race. In: European Conference on Computer Vision. pp. 344--360. Springer (2022)

\bibitem{marriott20213d}
Marriott, R.T., Romdhani, S., Chen, L.: A 3d gan for improved large-pose facial recognition. In: Proceedings of the IEEE/CVF Conference on Computer Vision and Pattern Recognition. pp. 13445--13455 (2021)

\bibitem{mehrabi2021survey}
Mehrabi, N., Morstatter, F., Saxena, N., Lerman, K., Galstyan, A.: A survey on bias and fairness in machine learning. ACM Computing Surveys (CSUR)  \textbf{54}(6),  1--35 (2021)

\bibitem{Moschoglou2020-ry}
Moschoglou, S., Ploumpis, S., Nicolaou, M.A., Papaioannou, A., Zafeiriou, S.: {3DFaceGAN}: Adversarial nets for {3D} face representation, generation, and translation. Int. J. Comput. Vis.  (May 2020)

\bibitem{ntoutsi2020bias}
Ntoutsi, E., Fafalios, P., Gadiraju, U., Iosifidis, V., Nejdl, W., Vidal, M.E., Ruggieri, S., Turini, F., Papadopoulos, S., Krasanakis, E., et~al.: Bias in data-driven artificial intelligence systems—an introductory survey. Wiley Interdisciplinary Reviews: Data Mining and Knowledge Discovery  \textbf{10}(3),  e1356 (2020)

\bibitem{ORTIGOSAHERNANDEZ201732}
Ortigosa-Hernández, J., Inza, I., Lozano, J.A.: Measuring the class-imbalance extent of multi-class problems. Pattern Recognition Letters  \textbf{98},  32--38 (2017)

\bibitem{AlgoFairness}
Pessach, D., Shmueli, E.: Algorithmic fairness. In: Machine Learning for Data Science Handbook: Data Mining and Knowledge Discovery Handbook, pp. 867--886. Springer (2023)

\bibitem{ramaswamy2021fair}
Ramaswamy, V.V., Kim, S.S., Russakovsky, O.: Fair attribute classification through latent space de-biasing. In: Proceedings of the IEEE/CVF conference on computer vision and pattern recognition. pp. 9301--9310 (2021)

\bibitem{Rothe-IJCV-2018}
Rothe, R., Timofte, R., Van~Gool, L.: Deep expectation of real and apparent age from a single image without facial landmarks. International Journal of Computer Vision  \textbf{126}(2-4),  144--157 (2018)

\bibitem{ruiz2020morphgan}
Ruiz, N., Theobald, B.J., Ranjan, A., Abdelaziz, A.H., Apostoloff, N.: Morphgan: One-shot face synthesis gan for detecting recognition bias. arXiv preprint arXiv:2012.05225  (2020)

\bibitem{Shen2018mf}
Shen, Y., Zhou, B., Luo, P., Tang, X.: {FaceFeat-GAN}: A two-stage approach for identity-preserving face synthesis  (2018)

\bibitem{tartaglione2021end}
Tartaglione, E., Barbano, C.A., Grangetto, M.: End: Entangling and disentangling deep representations for bias correction. In: Proceedings of the IEEE/CVF conference on computer vision and pattern recognition. pp. 13508--13517 (2021)

\bibitem{AgeGender}
Uchida, Y.: Age gender estimation: Keras implementation of a cnn network for age and gender estimation. (2021), \url{https://github.com/yu4u/age-gender-estimation}

\bibitem{REVISE}
Wang, A., Liu, A., Zhang, R., Kleiman, A., Kim, L., Zhao, D., Shirai, I., Narayanan, A., Russakovsky, O.: Revise: A tool for measuring and mitigating bias in visual datasets. International Journal of Computer Vision  \textbf{130}(7),  1790--1810 (2022)

\bibitem{Wu2019-ha}
Wu, Y., Yang, F., Xu, Y., Ling, H.: {Privacy-protective-GAN} for privacy preserving face {DE-identification}. J. Comput. Sci. Technol.  \textbf{34}(1),  47--60 (Jan 2019)

\bibitem{xu2018fairgan}
Xu, D., Yuan, S., Zhang, L., Wu, X.: Fairgan: Fairness-aware generative adversarial networks. In: 2018 IEEE International Conference on Big Data (Big Data). pp. 570--575. IEEE (2018)

\bibitem{antispoofingYang2022}
Yang, J., Lan, G., Xiao, S., Li, Y., Wen, J., Zhu, Y.: Enriching facial anti-spoofing datasets via an effective face swapping framework. Sensors (Basel)  \textbf{22}(13) (Jun 2022)

\bibitem{YE2019294}
Ye, L., Zhang, B., Yang, M., Lian, W.: Triple-translation gan with multi-layer sparse representation for face image synthesis. Neurocomputing  \textbf{358},  294--308 (2019)

\end{thebibliography}

\end{document}